\pdfoutput=1

\documentclass[11pt]{article}

\usepackage[]{ACL2023}

\usepackage{times}
\usepackage{latexsym}
\usepackage[utf8]{inputenc}
\usepackage{array}
\usepackage{graphicx}
\usepackage{xcolor}
\usepackage{tikz}
\usepackage{lineno}
\usepackage{multirow}

\usepackage[T1]{fontenc}

\usepackage[utf8]{inputenc}

\usepackage{microtype}
\usepackage{tikz}
\usepackage{forest}
\usepackage{xcolor}
\usepackage{hyperref}
\usepackage{adjustbox}
\usetikzlibrary{shadows, fit, positioning}

\usepackage{inconsolata}

\title{Conflicts in Texts: Data, Implications and Challenges}

\author{Siyi Liu \\
  University of Pennsylvania \\
  \texttt{siyiliu@seas.upenn.edu} \\\And
  Dan Roth \\
  University of Pennsylvania and Oracle AI\\
  \texttt{danroth@seas.upenn.edu} \\}

\begin{document}
\maketitle
\begin{abstract}

As NLP models become increasingly integrated into real-world applications, it becomes clear that there is a need to address the fact that models often rely on and generate conflicting information. Conflicts could reflect the complexity of situations, changes that need to be explained and dealt with, difficulties in data annotation, and mistakes in generated outputs. In all cases, disregarding the conflicts in data could result in undesired behaviors of models and undermine NLP models’ reliability and trustworthiness. This survey categorizes these conflicts into three key areas: (1) \textit{natural texts on the web}, where factual inconsistencies, subjective biases, and multiple perspectives introduce contradictions; (2) \textit{human-annotated data}, where annotator disagreements, mistakes, and societal biases impact model training; and (3) \textit{model interactions}, where hallucinations and knowledge conflicts emerge during deployment. While prior work has addressed some of these conflicts in isolation, we unify them under the broader concept of \textit{conflicting information}, analyze their implications, and discuss mitigation strategies. We highlight key challenges for developing conflict-aware and robust NLP systems, and propose concrete research directions to address them.

\end{abstract}

\section{Introduction}

The rapid advancement of natural language processing (NLP), particularly with the rise of large language models (LLMs), has led to their widespread adoption in daily tasks, information retrieval, and decision-making processes. However, the increasing complexity of these models reveals various types of conflicts at multiple stages, including training, annotation, and model interaction, affecting the reliability and trustworthiness of downstream applications. For example, training models on data containing factual contradictions, annotation disagreements, or prompts that contradict a model’s parametric knowledge can introduce inconsistencies with unpredictable consequences \cite{pavlick2019inherent, sap-etal-2019-risk}.

\begin{figure}
    \centering
    \includegraphics[width=\linewidth]{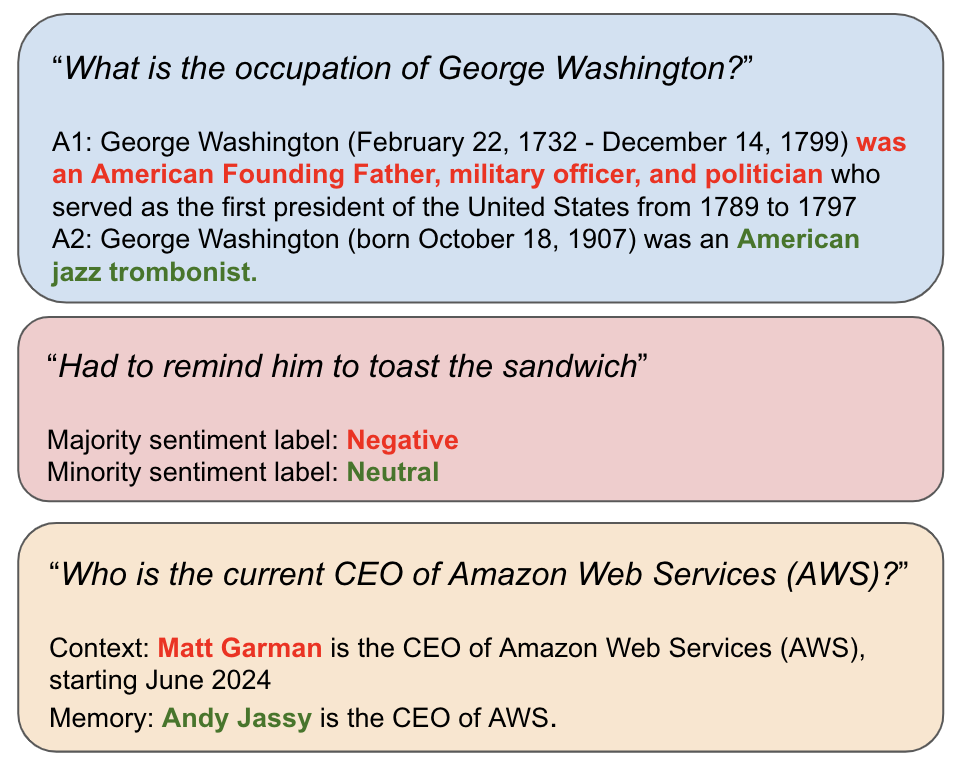}
    \caption{Examples of the three different areas of conflicts discussed in this work. The first example describes a case where two different entities of the same name are found \textit{naturally on the web}, the second example elaborates the \textit{annotation disagreement} in a sentiment analysis task, and the third showcases a knowledge conflict between the context and memory of LLMs during \textit{model interactions}.}
    \label{fig:enter-label}
\end{figure}

Existing work on conflicts in NLP tends to focus on specific issues, such as annotation disagreements~\cite{uma2021learning, klie2023analyzing}, hallucinations and factuality~\cite{zhang2023sirenssongaiocean, wang2023surveyfactualitylargelanguage}, and knowledge conflicts~\cite{xu-etal-2024-knowledge-conflicts, feng2024trendsintegrationknowledgelarge}, without synthesizing these problems into a broader perspective. In this survey, we conceptualize these diverse challenges under the umbrella of \textit{conflicting information} and analyze their origins, implications, and mitigation strategies. 

\begin{figure*}[t]
\begin{nolinenumbers}
    \centering
    \resizebox{\textwidth}{!}{
        \begin{forest}
        for tree={
            grow=east,
            parent anchor=east,
            child anchor=west,
            edge={draw, thick},
            rounded corners,
            draw=black,
            fit=band, 
            minimum width=5cm, 
            align=center, 
            anchor=west,
            s sep=18pt, 
            l sep=22pt, 
            inner sep=10pt, 
            font=\Large
        }
        [Conflicts in NLP, fill=yellow!40, tier=0
            [Model Interactions, fill=yellow!30, tier=1
                [Hallucination, fill=green!40, tier=2
                    [Contextual Hallucination, fill=red!40, tier=3
                        [{\citet{maynez-etal-2020-faithfulness}, \\
\citet{raunak-etal-2021-curious},
                \citet{dale-etal-2023-halomi}}, fill=cyan!40]
                    ]
                    [Factual Hallucination, fill=red!40, tier=3
                        [{\citet{lin2022truthfulqameasuringmodelsmimic}, \\
                \citet{pagnoni-etal-2021-understanding},
                    \citet{honovich-etal-2021-q2}}, fill=cyan!40]
                    ]   
                ]
                [Knowledge Conflicts, fill=green!40, tier=2
                    [Across Different Models, fill=red!40, tier=3
                        [{\citet{cohen2023lmvslmdetecting}, \\ 
\citet{zhu2024unravelingcrossmodalityknowledgeconflicts},
\citet{zhao2024knowingllmsknowsimple}}, fill=cyan!40]
                    ]
                    [Between Context and Parameters, fill=red!40, tier=3
                        [{\citet{longpre-etal-2021-knowledge, chen-etal-2022-rich},\\
                    \citet{chen2021dataset, lazaridou-etal-2021-mindthegap}}, fill=cyan!40]
                    ]
                ]
            ]
            [Human Annotations, fill=yellow!30, tier=1
                [Ethical and Societal Biases, fill=green!40, tier=2
                    [Annotation Biases, fill=red!40, tier=3
                        [{\citet{sap2022annotatorbias}, \citet{faisal-etal-2022-dataset}, \\
\citet{thorn-jakobsen-etal-2022-sensitivity} }, fill=cyan!40]
                    ]
                ]
                [Human Judgements, fill=green!40, tier=2
                    [Interpretation Disagreements, fill=red!40, tier=3
                        [{\citet{kahneman2021noise},
\citet{uma2021learning}, \\
\citet{sandri-etal-2023-dont},
\citet{jiang-marneffe-2022-investigating}}, fill=cyan!40]
                    ]
                ] 
            ]
            [Natural Conflicts on Web, fill=yellow!30, tier=1
                [Conflicts in Opinions, fill=green!40, tier=2
                    [Framing Bias, fill=red!40, tier=3
                        [{\citet{card-etal-2015-media},
\citet{liu-etal-2019-detecting}, \\
\citet{fan-etal-2019-plain},
\citet{lei-etal-2022-sentence}}, fill=cyan!40]
                    ]
                    [Perspectives, fill=red!40, tier=3
                        [{\citet{chen-etal-2019-seeing},
\citet{liu-etal-2021-multioped}, \\ \citet{lee-etal-2022-neus} ,
\citet{wan-etal-2024-convincing}
}, fill=cyan!40]
                    ]
                ]
                [Factual Conflicts, fill=green!40, tier=2
                    [Contradictory Evidence, fill=red!40, tier=3
                        [{\citet{chen-etal-2022-rich},\citet{hong2024gullible}, \\ \citet{liu2024opendomainquestionanswering}, \citet{pham2024whoswholargelanguage}}, fill=cyan!40]
                    ]
                    [Ambiguity, fill=red!40, tier=3
                        [{\citet{min-etal-2020-ambigqa}, \citet{zhang2021situatedqa}, \\ \citet{dhingra-etal-2022-time}, \citet{cole2023selective}}, fill=cyan!40]
                    ]
                ]
            ]
        ]
        \end{forest}
t    }
    
    \caption{Taxonomy of conflicts in texts.}
    \label{fig:conflicts_in_nlp}
    \end{nolinenumbers}
\end{figure*}

To ensure comprehensive and representative coverage of conflicts in NLP, we first establish a high-level categorization encompassing three primary sources: (1) natural conflicts present in web data, (2) conflicts arising from human annotation, and (3) conflicts emerging from model interactions. Notably, conflicts found in natural web texts and human-annotated datasets are primarily present in the training data—i.e., the inputs to models—whereas conflicts involving model interactions can arise in various forms, such as inconsistencies between model outputs and their inputs, contradictions among multiple outputs, or conflicts within the outputs themselves. For each category, we identify influential and widely cited survey papers as initial seed works ~\cite{uma2021learning, klie2023analyzing, zhang2023sirenssongaiocean, xu-etal-2024-knowledge-conflicts, feng2024trendsintegrationknowledgelarge, wang2023surveyfactualitylargelanguage}. Building upon these seeds, we systematically trace and incorporate the most impactful and representative studies for each type of conflict through citation chaining and targeted literature searches across major databases. This approach enables us to synthesize developments in each category and connect them, thereby providing an integrated discussion of current challenges, impacts on downstream tasks, and promising future directions for conflict-aware AI systems (See Appendix \ref{appendix:methodology} for more discussion about our survey methodology).

The abundance of \textbf{online data} is accompanied by inherent conflicts, stemming from diverse sources, interpretations, and biases. These conflicts manifest as \textit{factual conflicts}, such as semantic ambiguities~\cite{pavlick-tetreault-2016-empirical, min-etal-2020-ambigqa}\footnote{The previous version on ACL Anthology has a wrong paper title for a reference here.} and factual inconsistencies~\cite{pham2024whoswholargelanguage, liu2024opendomainquestionanswering}, or as \textit{conflicts in opinions} related to political ideologies~\cite{entman1993framing, recasens2013bias} and perspectives~\cite{chen-etal-2019-seeing, liu-etal-2021-multioped}. Factual conflicts are particularly prevalent in open-domain question answering (QA) and retrieval-augmented generation (RAG) systems \cite{chen2017reading}, where aggregating knowledge from multiple sources introduces inconsistencies \cite{liu2024opendomainquestionanswering}. These challenges highlight the need for conflict-aware retrieval and reasoning mechanisms to improve model reliability \cite{xie2024adaptive}. Unlike factual conflicts, opinionated disagreements reflect the variability in human interpretation, beliefs, and ideological stances \cite{chen-etal-2019-seeing, fan-etal-2019-plain}. The presence of conflicting viewpoints complicates tasks such as summarization, sentiment analysis, and dialogue generation, where maintaining coherence and neutrality is crucial \cite{liu-etal-2021-multioped, lee-etal-2022-neus}. Furthermore, the uneven distribution and biases of web data also affects models to behave from a Western perspective \cite{ramaswamy2023geodegeographicallydiverseevaluation, mihalcea2024ai}.

Another significant conflict arises in \textbf{human-annotated data}. For instance, \textit{annotation disagreements} persists in both subjective and seemingly objective NLP tasks~\cite{davani-etal-2022-dealing}. Disagreements are widespread in sentiment analysis~\cite{wan2023demographicdisagreement}, hate speech detection~\cite{sap2022annotatorbias}, and even natural language inference (NLI)~\cite{pavlick2019inherent}. Models trained on aggregated (e.g. majority-vote) labels struggle with ambiguous or high-disagreement examples, often treating them as hard-to-learn or mislabeled~\cite{anand2023subjective}. \citet{pavlick2019inherent} also find that standard NLI models' uncertainty does not reflect the true ambiguity present in human opinions, leading to overconfidence in contentious cases. In addition, \textit{annotation biases}—such as those related to race, gender, and geography—skew model predictions and reinforce societal biases~\cite{buolamwini2018gender, sap2022annotatorbias, pei-jurgens-2023-annotator}. These issues highlight the need for fair and representative annotations that capture the complexity of human disagreement.

Conflicts also emerge during \textbf{interactions with models}, manifesting as \textit{knowledge conflicts} between model memories and contexts, and \textit{hallucinations} in generated outputs. Knowledge conflicts arise when a model’s internal memory contradicts external contextual evidence, as shown by \citet{longpre-etal-2021-knowledge}, who found that models often overly depend on memorized knowledge, leading to hallucinations. 
\citet{neeman-etal-2023-disentqa} proposed separating parametric and contextual knowledge to improve interpretability, while \citet{xie2024adaptive} examined LLMs' confirmation bias, showing how models inconsistently handle contradictory evidence. 
Additionally, hallucinations—ranging from factual inconsistencies~\cite{lin2022truthfulqameasuringmodelsmimic, ouyang2022training} to contextual hallucinations~\cite{maynez-etal-2020-faithfulness, kryscinski-etal-2020-evaluating}—further undermine model reliability. Various mitigation strategies have been proposed, including retrieval augmentation~\cite{lewis-etal-2020-rag, shuster-etal-2021-retrieval}, hallucination detection~\cite{manakul-etal-2023-selfcheckgpt}, and knowledge graph-based verification~\cite{guan2024kgr}. 



In this survey, we systematically examine the landscape of conflicts in NLP by categorizing them into three primary sources.
For each conflict type, we detail how such conflicts arise and in what forms they take (\textbf{origins}), the challenges they pose (\textbf{implications}), and the strategies developed to address them (\textbf{mitigation}). We present a comprehensive taxonomy in Figure~\ref{fig:conflicts_in_nlp}, as well as structured summary tables—Table~\ref{tab:natural_texts_conflicts}, Table~\ref{tab:human_annotated_conflicts}, and Table~\ref{tab:model_interactions_conflicts}—that synthesize datasets, methodologies, and analysis from prior work. By offering a unified framework for understanding and addressing conflicting information in NLP, this survey contributes to the development of conflict-aware frameworks for data collection, model training, and model usage, ultimately enhancing the fairness and reliability of NLP.

\section{Conflicts in Natural Texts on the Web}
Conflicts in natural texts on the web manifest in diverse ways, reflecting the inherent complexity and subjectivity of human language. They can broadly be categorized into factual conflicts, which revolve around factual discrepancies caused by various reasons, and conflicts in opinions, which pertain to divergent perspectives or biases.

\subsection{Factual Conflicts}
\subsubsection{Origins}

\paragraph{Ambiguity}
Ambiguity is a root cause of factual conflict. When a query or piece of data lacks clarity about entities or context, a model can produce conflicting answers. A clear demonstration of how ambiguity induces conflicts is context dependence. For example, an ambiguous question of "which COVID-19 vaccine was the first to be authorized by our government?" can have conflicting answers depending on different geographical contexts \cite{zhang2021situatedqa}.

\citet{min-etal-2020-ambigqa} was the first work to study the effects of ambiguity in open domain question answering. They introduced AmbigQA, a dataset highlighting that over half of the open-domain, natural questions are ambiguous, with diverse sources of ambiguity such as event and entity references.  \citet{zhang2021situatedqa} proposed the SituatedQA task, showing that a significant fraction of open-domain questions are valid only under particular temporal or geographic contexts. Many other work specifically focus on the temporal aspect of ambiguity, benchmarking and evaluating models' awareness and adaptation to time-sensitive questions \cite{chen2021dataset, pmlr-v162-liska22a, NEURIPS2023_9941624e}.

\paragraph{Contradictory Evidence}
Conflicts in NLP systems arise when information on the web presents conflicting evidence towards a factual question. This issue is particularly prevalent in open-domain question answering settings, where models must navigate inconsistencies across diverse information sources. For example, \citet{liu2024opendomainquestionanswering} find that 25\% of unambiguous factual questions queried on Google retrieve conflicting evidence from multiple sources. 

Researchers have proposed different datasets to systematically study how NLP models handle such conflicts.
\citet{li-etal-2024-contradoc} introduce ContraDoc, a human-annotated dataset of long documents with internal contradictions; \citet{pham2024whoswholargelanguage} propose WhoQA, a benchmark dataset that constructs conflicts by formulating questions about a shared property among entities with the same name (e.g. "Who is George Washington?"); and \citet{liu2024opendomainquestionanswering} construct QACC, a human-annotated dataset of conflicting results retrieved by Google.
Beyond empirical datasets, several studies have proposed synthetic approaches to simulate conflicts through entity substitution \cite{chen-etal-2022-rich, hong2024gullible}, machine-generated conflicting evidence \cite{pan-etal-2023-misinformation, wan-etal-2024-convincing, hong2024gullible}, and pre-defined rule-based templates \cite{kazemi2023boardgameqa}.

\subsubsection{Implications and Mitigation}

\paragraph{Implications}
Factual conflicts pose significant challenges for NLP systems. Pre-trained language models accurately detect context-dependent questions but fall short when answering queries requiring temporal context, performing notably below human levels \cite{zhang2021situatedqa}. Additionally, large language models (LLMs) often exhibit confirmation bias, favoring retrieved information that aligns with their parametric memory despite contradictory evidence \cite{xie2024adaptive}. Consequently, conflicting information sources severely impact retrieval-augmented generation (RAG) frameworks, significantly degrading model performance even with minimal misinformation exposure \cite{pham2024whoswholargelanguage, liu2024opendomainquestionanswering, li-etal-2024-contradoc, pan-etal-2023-misinformation}.

\paragraph{Mitigation}
To address these challenges, various mitigation strategies have been proposed. Effective methods include fine-tuning calibrators for selective abstention \cite{chen-etal-2022-rich}, employing a "disambiguate-then-answer" pipeline to detect ambiguity proactively \cite{cole2023selective}, and developing time-aware models that condition responses on timestamps to manage outdated information \cite{dhingra-etal-2022-time}. Further robustness improvements have been achieved through fine-tuning discriminators or prompting GPT-3.5 models to explicitly recognize conflicting evidence \cite{hong2024gullible}, as well as incorporating human-written explanations in fine-tuning processes to enhance models' reasoning capabilities \cite{liu2024opendomainquestionanswering}.

\subsection{Conflicts in Opinions}

\subsubsection{Origins}
\paragraph{Perspectives}
Individuals and communities often hold diverse perspectives on the same issue. Such diversity is evident in online discussions and debates, where the multiplicity of viewpoints can lead to conflicting opinions. For instance, on controversial topics such as "Animals should have lawful rights," people express varying stances \cite{chen-etal-2019-seeing}, posing challenges for downstream tasks like summarization where consolidating viewpoints and presenting unbiased information are crucial \cite{liu-etal-2021-multioped, lee-etal-2022-neus}.

Several studies have explored perspectives in the context of conflicting information. \citet{chen-etal-2019-seeing} introduce the task of substantiated perspective discovery, where systems identify diverse, evidence-supported stances on a claim, and release the PERSPECTRUM dataset using online debates and search results. \citet{wan-etal-2024-convincing} propose ConflictingQA, a dataset of controversial questions paired with real-world documents that present divergent facts, arguments, and conclusions. \citet{plepi-etal-2024-perspective} examine perspective-taking in contentious online discourse, curating a corpus of 95k conflict scenarios annotated with users’ self-reported backgrounds. \citet{liu-etal-2021-multioped} present MultiOpEd, a corpus of 1,397 controversial topics, each paired with opposing editorials and concise summaries capturing their core perspectives.

\paragraph{Framing Bias}
A specific example of how differing opinions are conveyed and expanded is framing bias, a mechanism in which news media shape interpretations by emphasizing certain aspects of information over others \cite{entman1993framing}. In a polarized media environment, partisan media outlets deliberately frame news stories in a way to advance certain political ideologies \cite{jamieson2007effectiveness, levendusky2013partisan, liu-etal-2019-detecting}.

Numerous studies have investigated different aspects of media bias. \citet{card-etal-2015-media} introduce the Media Frames Corpus (MFC), a collection of news articles annotated with 15 general-purpose framing dimensions across three policy issues, enabling computational analysis of media framing. \citet{liu-etal-2019-detecting} present the Gun Violence Frame Corpus (GVFC), a dataset of news headlines annotated by domain experts to capture framing in gun violence reporting. \citet{fan-etal-2019-plain} examine informational bias—bias conveyed through content selection and structure—and release BASIL, a dataset of 300 news articles annotated with 1,727 bias spans, demonstrating that informational bias is more prevalent than lexical bias.

\subsubsection{Implications and Mitigation}

\paragraph{Implications} Analysis of PERSPECTRUM reveals significant natural language understanding challenges, as human performance substantially outperforms machine baselines at identifying diverse, evidence-supported perspectives \cite{chen-etal-2019-seeing}. Furthermore, when selecting real-world evidence for controversial questions, LLMs predominantly prioritize the relevance of the evidence to the query, often disregarding stylistic attributes such as the presence of scientific references or a neutral tone \cite{wan-etal-2024-convincing}. In addition, the distribution and biases of web data also affects models to behave from a Western perspective \cite{ramaswamy2023geodegeographicallydiverseevaluation, mihalcea2024ai}. 
Studies have shown that LLMs' outputs skew toward the values of Western English-speaking countries \cite{10.1093/pnasnexus/pgae346, naous-etal-2024-beer}, and misalignment is more pronounced for underrepresented personas and on culturally sensitive topics such as social values \cite{al-kuwatly-etal-2020-identifying}. Furthermore, LLMs often provide inconsistent answers to the same question when prompted in different languages \cite{li-etal-2024-land, alkhamissi-etal-2024-investigating, eloundou2025firstpersonfairnesschatbots}, revealing conflicting cultural perspectives within a single model.

\paragraph{Mitigation}

Several studies have proposed methods to address conflicts in perspectives and ideological bias. \citet{liu-etal-2021-multioped} show that auxiliary tasks improve perspective summarization quality, while \citet{chen-etal-2022-design} propose a retrieval paradigm that clusters documents by viewpoint, revealing users' preference for diverse perspectives over relevance-ranked lists. \citet{jiang-etal-2023-large} generate opinion summaries by selecting review subsets based on sentiment polarity and contrast, producing balanced pros, cons, and verdicts. \citet{plepi-etal-2024-perspective} demonstrate that conditioning generation on users' personal contexts yields more empathetic and appropriate responses than general-purpose models.

To mitigate framing and ideological bias, \citet{milbauer-etal-2021-aligning} uncover nuanced worldview differences across communities by identifying multiple axes of polarization beyond the traditional left–right spectrum. \citet{liu-etal-2022-politics} pre-train models for ideology detection by comparing reporting on the same events across partisan sources. \citet{chen-etal-2023-ideology} disentangle content from style to enable ideology classification under data scarcity and bias. \citet{lee-etal-2022-neus} employ hierarchical multi-task learning to neutralize bias from news titles to article bodies, while \citet{liu2023open} construct neutral event graphs by synthesizing perspectives across ideological divides.

\section{Conflicts in Human-Annotated Texts}

Conflicts in human-annotated texts largely arise from two sources: annotation disagreements and societal or ethical biases. Disagreements stem from linguistic ambiguity, annotator backgrounds, and task design, while biases reflect systematic demographic or ideological influences that can skew labeling in consistent ways. Though conceptually distinct, these sources often interact—biases may amplify disagreement or entrench disparities. Differentiating between them is essential for understanding annotation-related conflicts and for developing more reliable and equitable NLP datasets.
\subsection{Origins}

\paragraph{Annotation Disagreement}
The subjective nature of human judgment introduces variability and disagreement into annotated data~\cite{kahneman2021noise}. In NLP, such disagreements arise from linguistic ambiguity, annotator backgrounds, task design, and dataset curation practices. \citet{uma2021learning} survey disagreements across NLP and vision tasks, identifying subjective ambiguity and annotator diversity as key contributors. \citet{sandri-etal-2023-dont} classify disagreements in offensive language detection as stemming from inherent ambiguity, annotation errors, or contextual gaps, highlighting that some disagreements reflect hard-to-classify content, while others indicate correctable issues. Similarly, \citet{jiang-marneffe-2022-investigating} categorize NLI disagreements into linguistic uncertainty, annotator bias, and task design, showing that much of the observed noise is systematic and predictable.

Task formulation also plays a critical role. \citet{dsouza-kovatchev-2025-sources} find that label disagreement in reinforcement learning from human feedback (RLHF) is shaped by annotator selection and task phrasing. 
Demographic and ideological factors further influence disagreements. \citet{pavlick2019inherent} argue that many NLI disagreements reflect genuine linguistic ambiguity and individual variation rather than annotation error. \citet{sap2022annotatorbias} demonstrate that annotators' personal beliefs and identities affect toxicity judgments, while \citet{wan2023demographicdisagreement} show that demographic features significantly improve disagreement prediction.

\paragraph{Ethical and Societal Biases}
Human-annotated texts also encode societal biases related to race, gender, and geography, which can significantly skew model predictions and downstream decisions~\cite{buolamwini2018gender}. \citet{sap2022annotatorbias} show that annotators’ ideological and racial identities influence toxicity judgments, with conservative annotators less likely to flag anti-Black slurs and more likely to misclassify African American English (AAE) as offensive. \citet{thorn-jakobsen-etal-2022-sensitivity} examine how annotation guidelines interact with annotator demographics, demonstrating that even well-designed tasks can elicit systematically different responses across groups, highlighting the need for inclusive task design. \citet{pei-jurgens-2023-annotator} introduce POPQUORN, a dataset designed to assess demographic effects on annotation across NLP tasks, and find that annotator attributes—such as age, gender, race, and education—account for substantial variance in labeling behavior.

\subsubsection{Implications and Mitigation}
\paragraph{Implications}

Early research has underscored the impact of annotator disagreement on data quality and model performance~\cite{artstein-poesio-2008-survey, pustejovsky2012natural, plank2014learning}. \citet{pavlick2019inherent} show that standard NLI models fail to capture the true uncertainty present in human judgments, leading to overconfidence on contentious examples. Similarly, \citet{anand2023subjective} find that models trained on single “gold” labels perform poorly and exhibit lower confidence on high-disagreement instances, often treating them as mislabeled or hard to learn. \citet{sap-etal-2019-risk} demonstrate how annotator bias can yield discriminatory outcomes: tweets in African American English (AAE) are frequently misclassified as toxic, a bias inherited by models that disproportionately flag content from Black authors. Additionally, many widely used NLP datasets exhibit a strong Western-centric skew~\cite{faisal-etal-2022-dataset}, causing models to generalize poorly to underrepresented regions—for example, excelling on questions about New York or London, but failing on Nairobi or Manila due to lack of exposure.

\paragraph{Mitigation}
Prior work has explored collecting multiple labels per data item to capture annotation variability and improve data quality. Probabilistic models have been developed to infer true labels by accounting for annotator expertise and label noise~\cite{10.1145/1401890.1401965}. \citet{davani-etal-2022-dealing} propose a multi-task neural network that models each annotator’s labels individually while sharing a common representation, preserving disagreement in training. Similarly, studies show that models trained on soft labels—i.e., full label distributions reflecting annotator disagreement—consistently outperform those trained on aggregated labels~\cite{uma2021learning, fornaciari-etal-2021-beyond}.

\section{Conflicts during Model Interactions}

Conflicts during model interactions primarily manifest as knowledge conflicts and hallucinations, each posing distinct challenges. Knowledge conflicts occur when a model’s parametric memory contradicts contextual input or when inconsistencies arise across models, whereas hallucinations occur when outputs deviate from real-world facts or the given input. Differentiating these two types of conflict clarifies their underlying causes and helps guide targeted mitigation strategies.

\subsection{Knowledge Conflicts}
\subsubsection{Origins}

\paragraph{Context vs. Memory}

A common type of knowledge conflict arises when a model’s prompt (contextual knowledge) contradicts what the model has learned and stored in its parameters (parametric knowledge) \cite{longpre-etal-2021-knowledge, chen-etal-2022-rich}. One prevalent cause of such conflicts is the presence of updated information \cite{chen2021dataset, lazaridou-etal-2021-mindthegap, luu2022timewaitsoneanalysis}, where newly available knowledge contradicts models' previously learned knowledge.

Recent studies have developed many evaluation frameworks and datasets to assess LLMs' behaviors in this scenario through different methods, including entity substitution \cite{longpre-etal-2021-knowledge, chen-etal-2022-rich, wang2024resolvingknowledgeconflictslarge}, adversarial perturbation \cite{chen-etal-2022-rich, xie2024adaptive}, misinformation injection \cite{pan-etal-2023-misinformation}, and machine generation \cite{qian2023mergeconflictsexploringimpacts, ying-etal-2024-intuitive, tan-etal-2024-blinded}.

\paragraph{Within and Across Models}

Conflicts may also arise across or within model knowledge bases. \citet{cohen2023lmvslmdetecting} explore how different LLMs encode different knowledge and can be used to fact-check one another, uncovering inconsistencies indicative of factual errors. \citet{zhu2024unravelingcrossmodalityknowledgeconflicts} examine cross-modality conflicts in vision-language models, attributing discrepancies between visual and textual components to separate training regimes and distinct data sources. Even within a single model, contradictions can emerge: \citet{zhao2024knowingllmsknowsimple} detect intra-model inconsistencies by paraphrasing queries and observing divergent answers across prompts.


\subsubsection{Implications and Mitigation}
\paragraph{Implications}
Interestingly, different studies of knowledge conflicts present seemingly contradictory findings. Some studies claim that models often excessively rely on parametric memory when observing conflicts with contextual knowledge \cite{longpre-etal-2021-knowledge}; Some other studies posit that LLMs tend to ground their answers in retrieved documents in this scenario \cite{chen-etal-2022-rich, qian2023mergeconflictsexploringimpacts, tan-etal-2024-blinded}; or even both – LLMs are highly receptive to context when it is the only evidence presented in a coherent way, but also demonstrate a strong confirmation bias toward parametric memory when
both supportive and contradictory evidence to their parametric memory are present \cite{xie2024adaptive}.

\paragraph{Mitigation} 
Several approaches have been proposed to mitigate the impact of knowledge conflicts. \citet{longpre-etal-2021-knowledge} reduce memorization by augmenting training data through corpus substitution. \citet{chen-etal-2022-rich} introduce a calibrator that abstains from prediction when conflicting evidence is detected. More recently, \citet{wang2024resolvingknowledgeconflictslarge} propose an instruction-based framework that enables LLMs to identify conflicts, localize conflicting segments, and generate distinct responses for conflicting scenarios.


\subsection{Hallucination}
\subsubsection{Origins}
\paragraph{Factual Hallucinations}
Factual hallucinations arise when a model’s output contradicts real-world facts. 
\citet{lin2022truthfulqameasuringmodelsmimic} present TruthfulQA, an adversarial QA benchmark, and show that even top-performing models like GPT-3 were truthful on only 58\% of questions, compared to 94\% for humans. 
\citet{pagnoni-etal-2021-understanding} construct FRANK, a dataset for identifying factual errors in summarization, while \citet{honovich-etal-2021-q2} extend QAGS to dialogue by leveraging question generation and entailment for factual consistency evaluation. To assess factual knowledge and reasoning in LLMs, \citet{hu2023largelanguagemodelsknow} introduce Pinocchio, a large benchmark covering multiple domains, timelines, and languages, revealing challenges in composition, temporal reasoning, and robustness. \citet{mallen-etal-2023-trust} further find that models struggle with less common factual knowledge, with retrieval augmentation significantly improving performance in such cases.


\paragraph{Contextual Hallucinations}
Contextual hallucinations occur when generated text contradicts the given input context, such as in summarization, translation, and generation tasks. \citet{maynez-etal-2020-faithfulness} find that summarization models frequently generate content unfaithful to input documents, with 64\% of summaries containing unsupported information. In machine translation, \citet{raunak-etal-2021-curious} analyze hallucinations caused by source perturbations and training noise, and find that slight modifications to input data could trigger off-topic translations. Similarly, \citet{dale-etal-2023-halomi} introduce HalOmi, a multilingual benchmark for hallucination and omission detection in machine translation, showing that prior hallucination detectors often fail across different language pairs. In generation tasks, \citet{liu2021token} propose a novel token-level, reference-free hallucination detection task and dataset (HADES) for free-form text generation, and \citet{niu2024ragtruthhallucinationcorpusdeveloping} introduce RAGTruth, a comprehensive corpus designed for analyzing word-level hallucinations across various domains and tasks within standard Retrieval-Augmented Generation (RAG) frameworks.

\subsubsection{Implications and Mitigation}
\paragraph{Implications}
Hallucinations threaten trust, safety, and the integrity of AI-powered workflows. Hallucinated outputs can rapidly spread false information. For instance, in 2023, an AI-generated image purporting to show an explosion near the Pentagon went viral, briefly causing public panic and even a stock market dip before being debunked \cite{sun2024ai}. Hallucinations directly degrade the performance of downstream applications like abstractive summarization. Studies have found that a large portion of generated summaries contain unsupported facts, misleading readers and propagating misinformation in news and scientific dissemination \cite{kryscinski-etal-2020-evaluating}.

\paragraph{Mitigation}
Mitigating hallucinations in language models has been approached through various strategies, including knowledge disentanglement \cite{neeman-etal-2023-disentqa}, retrieval augmentation \cite{lewis-etal-2020-rag, shuster-etal-2021-retrieval}, knowledge graphs \cite{guan2024kgr}, and improved verification methods \cite{kryscinski-etal-2020-evaluating, wang-etal-2020-asking, laban2022summaC, manakul-etal-2023-selfcheckgpt}. DisentQA enhances robustness by training models to separate internal memory from external context, improving accuracy in conflicting knowledge scenarios \cite{neeman-etal-2023-disentqa}. Retrieval-Augmented Generation (RAG) mitigates factual inconsistencies by integrating external sources like Wikipedia \cite{lewis-etal-2020-rag} or incorporating a neural search module into chatbot responses \cite{shuster-etal-2021-retrieval}. In addition, \citet{guan2024kgr} demonstrate how retrofitting LLM outputs using structured knowledge graphs can correct factual inconsistencies, particularly in complex reasoning tasks. For hallucination detection methods, FactCC and QAGS introduce automated methods using synthetic data and question-answer validation to assess factual consistency \cite{kryscinski-etal-2020-evaluating, wang-etal-2020-asking}. SummaC refines entailment-based scoring \cite{laban2022summaC}, and SelfCheckGPT detects hallucinations by sampling multiple model outputs and checking for agreement without external references \cite{manakul-etal-2023-selfcheckgpt}.

\section{Connections, Challenges and Directions}
Given the significance and impact of conflicts in NLP, we advocate for increased attention to the development of conflict-aware and robust AI systems. In this section, we highlight specific challenges by connecting different types of conflicts and propose concrete research directions to address them.

\paragraph{Culturally Robust LLMs}
Among the challenges outlined in this survey, the development of culturally robust LLMs remains particularly underexplored. Cultural conflicts emerge both in naturally occurring web data and human-annotated datasets, where Western-centric distributions dominate. Prior studies reveal that LLMs often reflect the values and perspectives of Western, English-speaking populations~\cite{ramaswamy2023geodegeographicallydiverseevaluation, mihalcea2024ai, 10.1093/pnasnexus/pgae346, naous-etal-2024-beer}, with misalignments especially pronounced for underrepresented personas and culturally sensitive topics~\cite{al-kuwatly-etal-2020-identifying}. Additionally, LLMs exhibit inconsistent behavior across languages~\cite{li-etal-2024-land, alkhamissi-etal-2024-investigating, eloundou2025firstpersonfairnesschatbots}, revealing internal cultural conflicts. These issues are rooted in the data: both the pre-train data and benchmark datasets commonly exhibit Western-centric biases~\cite{mihalcea2024ai,faisal-etal-2022-dataset}, causing models to default to Western contexts and perform poorly on less-represented regions and cultures.

However, to the best of our knowledge, no effective methodology has yet been proposed to address this issue. With the emergence of culturally distinct LLMs—such as Qwen, trained largely on Chinese data~\cite{bai2023qwentechnicalreport}, and Vikhr, trained on Russian data~\cite{nikolich2025vikhrfamilyopensourceinstructiontuned}—a promising direction is model fusion across culturally diverse models to achieve greater cultural balance~\cite{wan2024knowledge, jiang-etal-2023-llm}. Furthermore, advances in culture-specific LLMs and synthetic data generation offer the potential to curate more culturally representative training and evaluation datasets beyond Western-centric narratives, supporting the development of culturally robust LLMs.

\paragraph{Building Conflict-Aware AI Systems}

As outlined in this survey, various types of conflicts can arise in a model’s input, each requiring different handling depending on the task. We argue that downstream applications should not treat all conflicts uniformly; rather, responses should be tailored to the conflict type. For instance, conflicts due to ambiguity should elicit clarification questions, factual contradictions should trigger reasoning over evidence, and opinion-based disagreements should induce balanced, multi-perspective responses. Realizing such capabilities requires models to be aware of the potential conflicts and classify them according to a systematic taxonomy. Yet, current research lacks frameworks to distinguish and operationalize these conflict types. Our proposed taxonomy offers a foundational step toward enabling conflict-aware systems that can recognize, interpret, and appropriately address diverse conflicts in downstream applications.

\section{Conclusion}
We present a unified view of \emph{conflicting information} in NLP, organizing the landscape into conflicts originating from (i) natural texts on the web, (ii) human annotations, and (iii) model interactions. This taxonomy connects lines of work that are often studied in isolation and clarifies how conflicts arise, what they imply for reliability, and how current methods aim to mitigate them. Our synthesis argues that conflict awareness should guide the full pipeline: data collection that preserves disagreement and multiple perspectives, models that detect and categorize conflicts and respond with clarification, reasoning, or balanced presentation, and evaluations that measure calibration under disagreement, robustness to contradictory evidence, and cultural coverage across languages and regions.

\section*{Limitations}
Conflicting information is present both in the data that models rely on and in their generated outputs. While we strive to account for all potential conflict scenarios, some cases may inevitably be overlooked. Additionally, due to space constraints, we cannot provide an exhaustive discussion of the literature on each specific type of conflict. Instead, we adopt a broader perspective, examining various types of conflicts to identify connections, challenges, and future directions.

\section*{Acknowledgment}
This research is based upon work supported in part by the Office of the Director of National Intelligence (ODNI), Intelligence Advanced Research Projects Activity (IARPA), via 2022-22072200003. The views and conclusions contained herein are those of the authors and should not be interpreted as necessarily representing the official policies, either expressed or implied, of ODNI, IARPA, or the U.S. Government. The U.S. Government is authorized to reproduce and distribute reprints for governmental purposes notwithstanding any copyright annotation therein. This work is also supported by NSF grant \#IIS-2135581.

\bibliography{custom}
\bibliographystyle{acl_natbib}

\appendix
\newpage

\section{Summary Tables}
In the summary tables, \textit{dataset} covers prior work that proposed datasets and benchmark, \textit{method} covers work that focus on mitigation strategies, and \textit{analysis} presents work that aim at providing insights through experiments.

\begin{table*}[t]
\centering
\caption{Datasets, methods, and analysis for conflicts in natural texts}
\label{tab:natural_texts_conflicts}
\begin{tabular}{|c|c|c|l|}
\hline
\textbf{Conflict Type}       & \textbf{Sub-type}         & \textbf{Category} & \textbf{Work}         \\
\hline
\multirow{20}{*}{Factual}
    & \multirow{7}{*}{Ambiguity}
        & \multirow{5}{*}{Dataset} & SituatedQA \cite{zhang2021situatedqa} \\
    &   &  & AmbigQA \cite{min-etal-2020-ambigqa} \\
    &   &  & Time-sensitive QA \cite{chen2021dataset} \\
    &   &  & StreamingQA \cite{pmlr-v162-liska22a} \\
    &   &  & Real-time QA \cite{NEURIPS2023_9941624e} \\
    \cline{3-4}
    &                           & \multirow{2}{*}{Method}  & Disambiguate then answer \cite{cole2023selective}        \\
    &                           &                          & Time-aware LM  \cite{dhingra-etal-2022-time}     \\
    \cline{2-4}
    & \multirow{13}{*}{Contradictory Evidence}
        & \multirow{9}{*}{Dataset} & QACC \cite{liu2024opendomainquestionanswering}      \\
    &   &  & Contra-Doc \cite{li-etal-2024-contradoc} \\
    &   &  & WhoQA \cite{pham2024whoswholargelanguage} \\
    &   &  & Machine-generated \cite{pan-etal-2023-misinformation} \\
    &   &  & Machine-generated \cite{wan-etal-2024-convincing} \\
    &   &  & Machine-generated \cite{hong2024gullible} \\
    &   &  & Machine-generated \cite{jiayang-etal-2024-econ} \\
    &   &  & Entity-substitution \cite{chen-etal-2022-rich} \\
    &   &  & Rule-based \cite{kazemi2023boardgameqa} \\
    \cline{3-4}
    &                           & \multirow{3}{*}{Method}  & Finetuned Calibrator \cite{chen-etal-2022-rich}       \\
    &                           &                          & Finetuned w/ Explanation \cite{liu2024opendomainquestionanswering}       \\
    &                           &                          & Finetuned discriminator \cite{hong2024gullible}       \\
    \cline{3-4}
    &                           & \multirow{1}{*}{Analysis}  & Confirmation bias \cite{xie2024adaptive}       \\
\hline

\multirow{16}{*}{Opinion }
    & \multirow{8}{*}{Perspectives}
        & \multirow{5}{*}{Dataset} & PERSPECTRUM \cite{chen-etal-2019-seeing}       \\
    &                           &                          & Multi-OpEd \cite{liu-etal-2021-multioped}       \\
    &                           &                          & NeuS \cite{lee-etal-2022-neus}       \\
    &                           &                          & ConflictingQA \cite{wan-etal-2024-convincing}       \\
    
    &                           &                          &  Reddit \cite{plepi-etal-2024-perspective}       \\
    \cline{3-4}
    &                           & \multirow{3}{*}{Method}  & Multi-task learning \cite{liu-etal-2021-multioped}        \\
    &                           &                          & Opinion summarization \cite{jiang-etal-2023-large}        \\
    &                           &                          & Tailored generation \cite{plepi-etal-2024-perspective}        \\
    \cline{2-4}
    & \multirow{7}{*}{Framing Bias}
        & \multirow{3}{*}{Dataset} & MFC \cite{card-etal-2015-media}      \\
    &                           &                          & GVFC \cite{liu-etal-2019-detecting}      \\
    &                           &                          & BASIL \cite{fan-etal-2019-plain}      \\
    
    \cline{3-4}
    &                           & \multirow{4}{*}{Method}  &   Multifaceted analysis \cite{milbauer-etal-2021-aligning}     \\
    &                           &                          & Pre-training \cite{liu-etal-2022-politics}       \\
    &                           &                          & Disentanglement \cite{chen-etal-2023-ideology}   \\
    &                           & 
    & Multi-task learning \cite{lee-etal-2022-neus}   \\
    \cline{3-4}
    &                           & \multirow{1}{*}{Analysis}  & Sentence-level \cite{lei-etal-2022-sentence} \\
\hline
\end{tabular}
\end{table*}

\begin{table*}[t]
\centering
\caption{Datasets, methods, and analysis for conflicts in human-annotated texts}
\label{tab:human_annotated_conflicts}
\begin{tabular}{|c|c|c|l|}
\hline
\textbf{Conflict Type}      & \textbf{Sub-type}            & \textbf{Category} & \textbf{Work}                                   \\
\hline
\multirow{22}{*}{Human-Annotated}
  & \multirow{16}{*}{Disagreement}
    & \multirow{4}{*}{Dataset}
      & Twitter~\cite{sandri-etal-2023-dont}             \\
  &                                &                      & RLHF~\cite{dsouza-kovatchev-2025-sources}   \\
  &                                &                      & DiscoGeM~\cite{yung-demberg-2025-crowdsourcing}       \\
  &                                &                      & NLI~\cite{pavlick2019inherent}     \\
  \cline{3-4}
  &                                & \multirow{4}{*}{Method}
      & Probabilistic model~\cite{10.1145/1401890.1401965}    \\
  &                                &                      & Multi-task~\cite{davani-etal-2022-dealing}    \\
  &                                &                      & Soft labels~\cite{uma2021learning}               \\
  &                                &                      & Soft labels~\cite{fornaciari-etal-2021-beyond} \\
   \cline{3-4}
    &                           & \multirow{8}{*}{Analysis}  & Survey \cite{uma2021learning} \\
    &                                &                      & Survey~\cite{klie2023analyzing} \\
    &                                &                      & Offensive language~\cite{sandri-etal-2023-dont} \\
    &                                &                      & NLI~\cite{jiang-marneffe-2022-investigating} \\
     &                                &                      & Task design~\cite{dsouza-kovatchev-2025-sources} \\
     &                                &                      & Free choice~\cite{yung-demberg-2025-crowdsourcing} \\
     &                                &                      & Personal belief~\cite{sap2022annotatorbias} \\
     &                                &                      & Demographic data~\cite{wan2023demographicdisagreement} \\
  \cline{2-4}
  & \multirow{6}{*}{Biases}
    & \multirow{3}{*}{Dataset}
      & Gender~\cite{buolamwini2018gender}         \\
  &                                &                      & Argument mining \cite{thorn-jakobsen-etal-2022-sensitivity} \\
  &                                &                      & POPQUORN~\cite{pei-jurgens-2023-annotator}       \\
  \cline{3-4}
  &                           & \multirow{3}{*}{Analysis}  & Western-centric \cite{faisal-etal-2022-dataset} \\
  &                                &                      &  Toxicity~\cite{wan2023demographicdisagreement} \\
  &                                &                      &  Racist outcome~\cite{sap-etal-2019-risk} \\
  
  \cline{3-4}
\hline
\end{tabular}
\end{table*}

\begin{table*}[t]
\centering
\caption{Datasets, methods, and analysis for conflicts during model interactions}
\label{tab:model_interactions_conflicts}
\begin{tabular}{|c|c|c|l|}
\hline
\textbf{Conflict Type} & \textbf{Sub-type}             & \textbf{Category} & \textbf{Work}                                                        \\
\hline
\multirow{12}{*}{Knowledge}
  & \multirow{9}{*}{Context vs. Memory}
      & \multirow{7}{*}{Dataset}
          & Entity substitution \cite{longpre-etal-2021-knowledge}    \\
  &                           &                         &Entity substitution \cite{chen-etal-2022-rich}                   \\
  &                           &                         & Instruction-based \cite{wang2024resolvingknowledgeconflictslarge} \\
  &                           &                         & Misinformation injection \cite{pan-etal-2023-misinformation}  \\
  &                           &                         & KRE \cite{ying-etal-2024-intuitive}         \\
  &                           &                         &  context-conflicting \cite{tan-etal-2024-blinded}             \\
  \cline{3-4}
  &                           & \multirow{3}{*}{Method}
          & Data Augmentation \cite{longpre-etal-2021-knowledge}         \\
  &                           &                         & Abstention \cite{chen-etal-2022-rich}                  \\
  &                           &                         & Instruction-based \cite{wang2024resolvingknowledgeconflictslarge}                  \\
  
  \cline{2-4}
  & \multirow{3}{*}{Within \& Across}
      & \multirow{3}{*}{Analysis}
          & LM-vs-LM fact-checking \cite{cohen2023lmvslmdetecting}             \\
  &                           &                         & Cross-modality  \cite{zhu2024unravelingcrossmodalityknowledgeconflicts} \\
  &                           &                         & Intra-model contradiction \cite{zhao2024knowingllmsknowsimple} \\
\hline
\multirow{23}{*}{Hallucination}
  & \multirow{14}{*}{Factual}
      & \multirow{5}{*}{Dataset}
          & TruthfulQA \cite{lin2022truthfulqameasuringmodelsmimic}            \\
  &                           &                         & FRANK \cite{pagnoni-etal-2021-understanding}   \\
  &                           &                         & $q^2$ \cite{honovich-etal-2021-q2}   \\
  &                           &                         & Pinocchio  \cite{hu2023largelanguagemodelsknow}          \\
  &                           &                         & MiniCheck  \cite{tang-etal-2024-minicheck}          \\
  \cline{3-4}
  &                           & \multirow{8}{*}{Method}
          & RAG \cite{lewis-etal-2020-rag}     \\
          &                           &                         & RAG  \cite{shuster-etal-2021-retrieval}          \\
          &                           &                         & Knowledge graph  \cite{guan2024kgr}          \\
          &                           &                         & Disentanglement  \cite{neeman-etal-2023-disentqa}          \\
          &                           &                         & QA validation  \cite{kryscinski-etal-2020-evaluating}          \\
          &                           &                         & QA validation  \cite{wang-etal-2020-asking}          \\
          &                           &                         & Entailment-based  \cite{laban2022summaC}          \\
          
          &                           &                         & SelfCheckGPT  \cite{manakul-etal-2023-selfcheckgpt}          \\
          \cline{3-4}
  &                           & \multirow{1}{*}{Analysis}
          & Less popular entities \cite{mallen-etal-2023-trust}         \\
  \cline{2-4}
  & \multirow{9}{*}{Contextual}
      & \multirow{3}{*}{Dataset}
          & HalOmi \cite{dale-etal-2023-halomi}           \\
  &                           &                         & HADES \cite{liu2021token}                       \\
  &                           &                         & RAGTruth \cite{niu2024ragtruthhallucinationcorpusdeveloping}\\
  \cline{3-4}
  &                           & \multirow{4}{*}{Method}
          & Context-aware decoding \cite{shi-etal-2024-trusting}\\
          &                           &                         & Long context \cite{liu-etal-2025-towards}  \\
  &                           &                         & Context-DPO \cite{bi2024contextdpoaligninglanguagemodels}  \\
  &                           &                         & CR-DPO \cite{huang2025to}  \\
\cline{3-4}
  &                           & \multirow{2}{*}{Analysis}
          & Summarization  \cite{maynez-etal-2020-faithfulness}\\
            &                           &                         & Translation   \cite{raunak-etal-2021-curious}  \\

\hline
\end{tabular}
\end{table*}

\section{Survey Methodology}
\label{appendix:methodology}
Our objective is to map the broad and diverse landscape of textual conflicts in NLP, identify central implications and gaps, and outline a research agenda. We surveyed research indexed in Google Scholar, the ACL Anthology, and OpenReview, covering peer-reviewed work in major ML and NLP venues such as ICLR and ACL, as well as recent manuscripts on arXiv. Rather than relying on predetermined keyword filters, we primarily used citation chaining: starting from influential surveys corresponding to the three sources of conflict considered in this work (natural web text, human annotations, and model interactions), we applied both backward and forward chaining using Google Scholar and venue indexes. Paper selection did not follow a rigid checklist. The two authors independently screened and included papers based on inclusion criteria that prioritized recency, citation impact, and perceived influence on the area, then reconciled disagreements through discussion. This methodology emphasizes coverage and structure over exhaustiveness and is intended to synthesize a rapidly evolving field while making the scope of inclusion explicit and surfacing open problems.

\end{document}